\documentclass[10pt,twocolumn,letterpaper]{article}

\usepackage{cvpr}              

\usepackage{graphicx}
\usepackage[accsupp]{axessibility}  
\usepackage{amsmath}
\usepackage{amssymb}
\usepackage{booktabs}
\usepackage{adjustbox}
\usepackage{ tipa }
\usepackage{multirow} 
\usepackage{mathtools}

\usepackage[pagebackref,breaklinks,colorlinks]{hyperref}
\newcommand{\squeezeup}{\vspace{-2mm}}

\usepackage[capitalize]{cleveref}
\crefname{section}{Sec.}{Secs.}
\Crefname{section}{Section}{Sections}
\Crefname{table}{Table}{Tables}
\crefname{table}{Tab.}{Tabs.}

\def\cvprPaperID{14} 
\def\confName{CVPR WMF}
\def\confYear{2023}

\begin{document}

\title{EKILA: Synthetic Media Provenance and Attribution for Generative Art}

\author{Kar Balan$^1$, Shruti Agarwal$^2$, Simon Jenni$^2$, Andy Parsons$^{2}$, Andrew Gilbert$^{1}$, John Collomosse$^{1,2}$\\
$^1$University of Surrey, $^2$Adobe Inc.\\
{\tt\small \{k.balan, a.gilbert\}@surrey.ac.uk},~~
{\tt\small \{shragarw, jenni, andyp, collomos\}@adobe.com}}

\twocolumn[{%
\renewcommand\twocolumn[1][]{#1}%
\maketitle
\vspace{-25pt}
\begin{center}
    \centering
    \includegraphics[width=\linewidth]{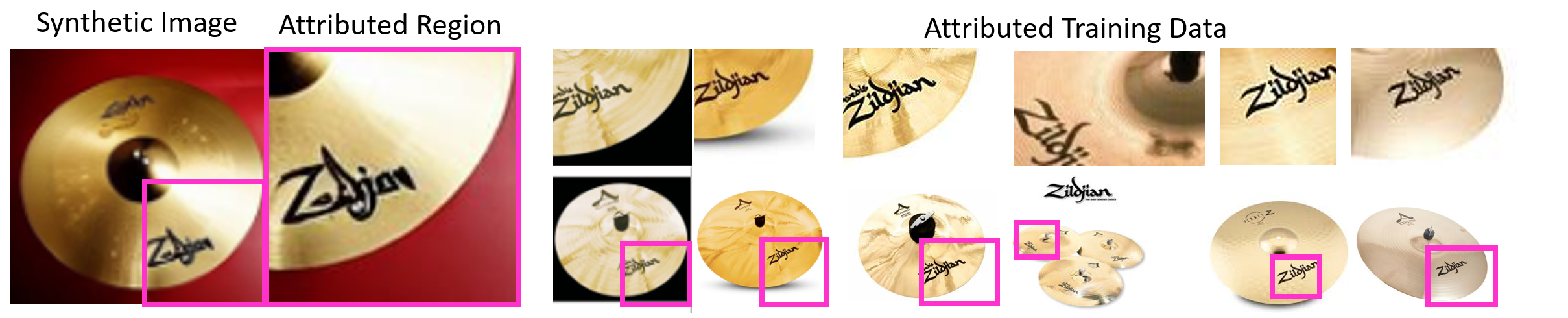}

    \label{fig:teaser}
Figure 1.  EKILA combines robust visual attribution with Distributed Ledger Technology (DLT) to recognize and reward creative contributions to generative art.  A cymbal generated by a Latent Diffusion Model (LDM) trained on LAION-400M is attributed to a subset of training images, credit weight apportioned, and royalties paid using our proposed method.

\end{center}%

 \vspace{10pt}
 
}]
\setcounter{figure}{1}

\begin{abstract}

         We present EKILA; a decentralized framework that enables creatives to receive recognition and reward for their contributions to generative AI (GenAI).  EKILA proposes a robust visual attribution technique and combines this with an emerging content provenance standard (C2PA) to address the problem of synthetic image provenance -- determining the generative model and training data responsible for an AI-generated image.  Furthermore, EKILA extends the non-fungible token (NFT) ecosystem to introduce a tokenized representation for rights, enabling a triangular relationship between the asset's Ownership, Rights, and Attribution (ORA). Leveraging the ORA relationship enables creators to express agency over training consent and, through our attribution model, to receive apportioned credit, including royalty payments for the use of their assets in GenAI.

\end{abstract}

\section{Introduction}
\label{sec:intro}

Generative AI (GenAI) is transforming digital art, creating compelling synthetic images by sampling millions of diverse creative works \cite{ldm,imagen,dalle2}. However, describing the provenance of synthetic assets -- \ie their generative models and the assets used to train them -- is critical to the acceptance of GenAI.  Much as the music industry matured from ad-hoc sampling in the eighties to a formal model for consent and paid reuse, the ability to commodify the image data sampled to train GenAI models may establish a new facet of our future creative economy.

To this end, we propose EKILA\footnote{From eh$\cdot$ki$\cdot$l\textschwa: ``complex rules that help people understand how to share in the right way." o. Mbendjele, D.C. Congo.\cite{ekila}}; a decentralized framework for assigning usage rights to creative assets and compensating creators when those rights are exercised. We tie the latter to a novel attribution and apportionment method for establishing synthetic images' provenance (origins), providing a means for creatives to be recognized and rewarded for their contributions to GenAI.  

EKILA is decentralized in that it uses tokenized representations of creative assets (Non-fungible Tokens: `NFTs').  NFTs are immutable digital representations of creative works created (`minted') on the blockchain \cite{erc721}.  One problem with NFTs in their current form is that purchasers are not conveyed any concrete `rights’ (\eg to make derivative works via GenAI or otherwise) \cite{nftlaw}.   This limits the scope for NFT use in creative works, leaving their primary use as a vehicle for financial speculation \cite{nftresell}.  Furthermore, {\em ownership provenance} and {\em creation provenance} are not coupled in NFTs, leaving assets open to unauthorized re-sale (`copy-minting' \cite{nftsurvey}). EKILA addresses these shortcomings by extending NFTs from the consideration solely of {\em ownership provenance} to the provenance of any {\em usage right} related to the NFT and links this to the {\em creation} provenance of the asset using an emerging open standard (the Coalition for Content Provenance and Authenticity: `C2PA' \cite{c2pa}).  As such, EKILA provides a means for describing the ownership, rights, and attribution of assets without recourse to a centralized registry, leveraging this in the context of GenAI to enable fair recognition and reward for creators.  We make three technical contributions:

{\textbf{1. ORA Triangle}.  We extend NFTs to propose a model that jointly describes the (O)wnership, usage, (R)ights, and (A)ttribution of assets.  ORA assets are `owned' by a smart contract operated by the creator rather than directly by an individual assignee, enabling tokenized rights to be assigned in a one-many relationship. Under ORA, NFT assets also bear attribution metadata (C2PA) describing their creation provenance.  Rights tokens are tied to this C2PA metadata to prevent substitution of the asset. This three-way binding (the `ORA Triangle') addresses two major limitations of NFTs: dynamic assignment of image rights and protection against `copy-minting' \cite{nftsurvey}.  

 {\textbf{2. Synthetic Media Attribution.}  We apply the C2PA provenance standard \cite{c2pa} to describe synthetic image provenance, namely the GenAI model that generated an image and the training data of that model.  We propose a robust visual matching method to attribute synthetic images (or parts thereof) to a subset of that training data and to apportion credit to recognize the creative contribution of training data responsible for that image.  We show this method to operate robustly over millions of images and image patches and, notably, to outperform CLIP \cite{clip} and common perceptual metrics (LPIPS \cite{lpips}, SIFID \cite{sifid}) at the attribution task.}

{\textbf{3. Royalty mechanism.} We combine both (1) and (2) to automatically compensate creators via crypto-currency payment for their contributions to synthetic art.  The royalty mechanism enables creators to receive apportioned credit via crypto-currency payment when their training images are attributed to generated images.


\section{Related Work}

\textbf{Diffusion models} underpin recent advances in GenAI  \cite{sohl2015deep,dhariwal2021diffusion,ho2020denoising,song2020denoising}: DALL-E 2 \cite{dalle2}, Imagen \cite{imagen}, and Stable Diffusion \cite{ldm} require hundreds of millions of images to train and improve recent large GAN models\cite{karras2019style,karras2020analyzing} in terms of quality and diversity.  Diffusion models typically condition generation upon a text-based prompt \eg, encoded via CLIP \cite{radford2021learning} often with additional modalities for fine-grained control of visual attributes \cite{li2020image,gafni2022make,pavllo2020controlling}.  Recent studies have shown that duplication in training data can lead to content \cite {carlini2023} or style memorization \cite{somepalli2022} in diffusion models. Inversion attacks seek to hallucinate representative examples for given classes or prompts \cite{fredikson2015,pavllo2020controlling} often resembling training data.  The problem of determining training set membership for diffusion models \cite{wu2022,hu2023} has also been studied.   Our work is closest to visual attribution approaches using CLIP \cite{somepalli2022,stableattribution} that measure the semantic similarity between generated and training whole images. However, we match on style and ``patchified" local structure proposing a new model for this purpose which we show to outperform semantic \cite{somepalli2022} and RGB patch correlations \cite{carlini2023}.

\begin{figure*}[t!]
    \centering
    \includegraphics[width=0.35\linewidth]{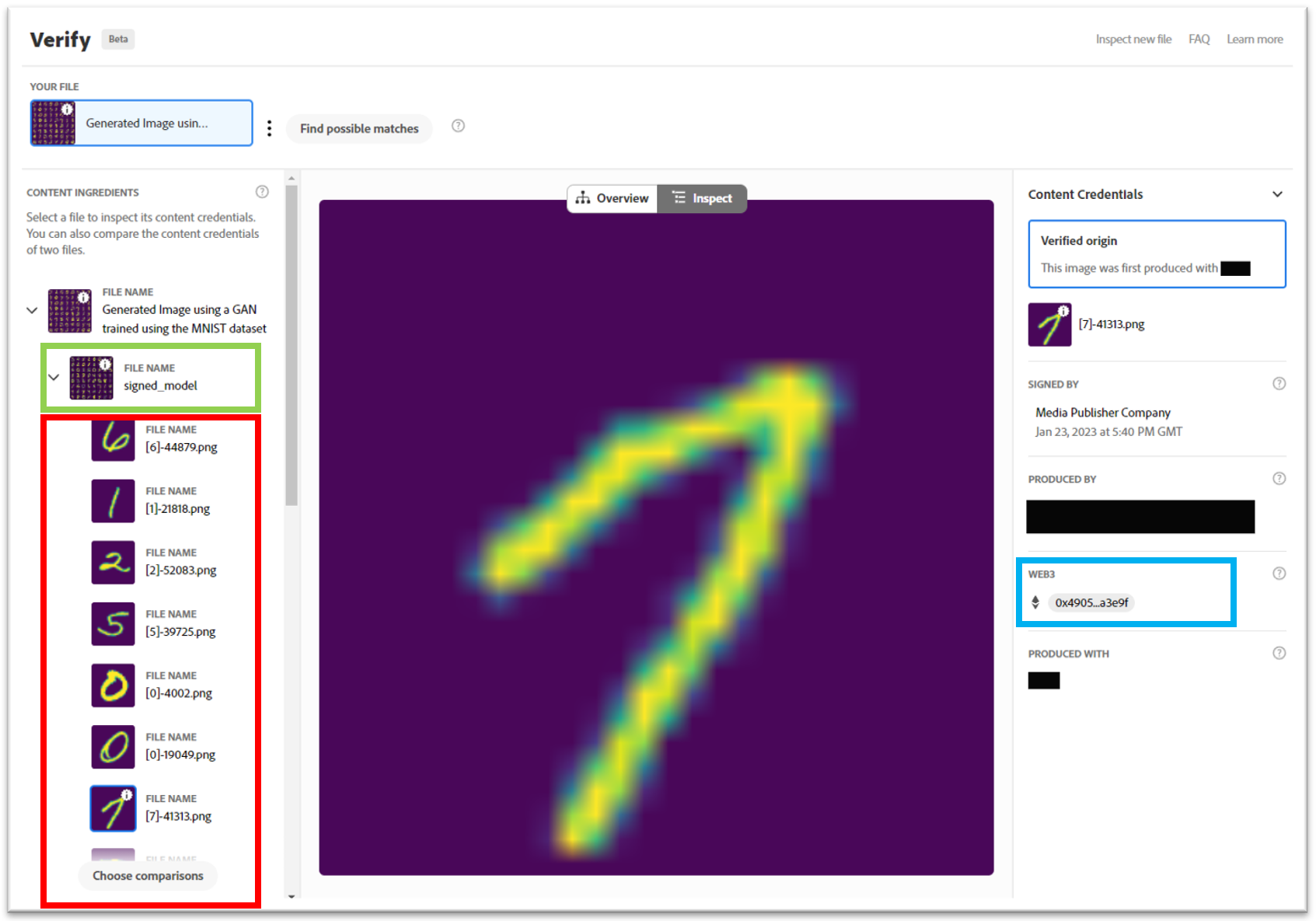}~~~~
    \includegraphics[width=0.62\linewidth]{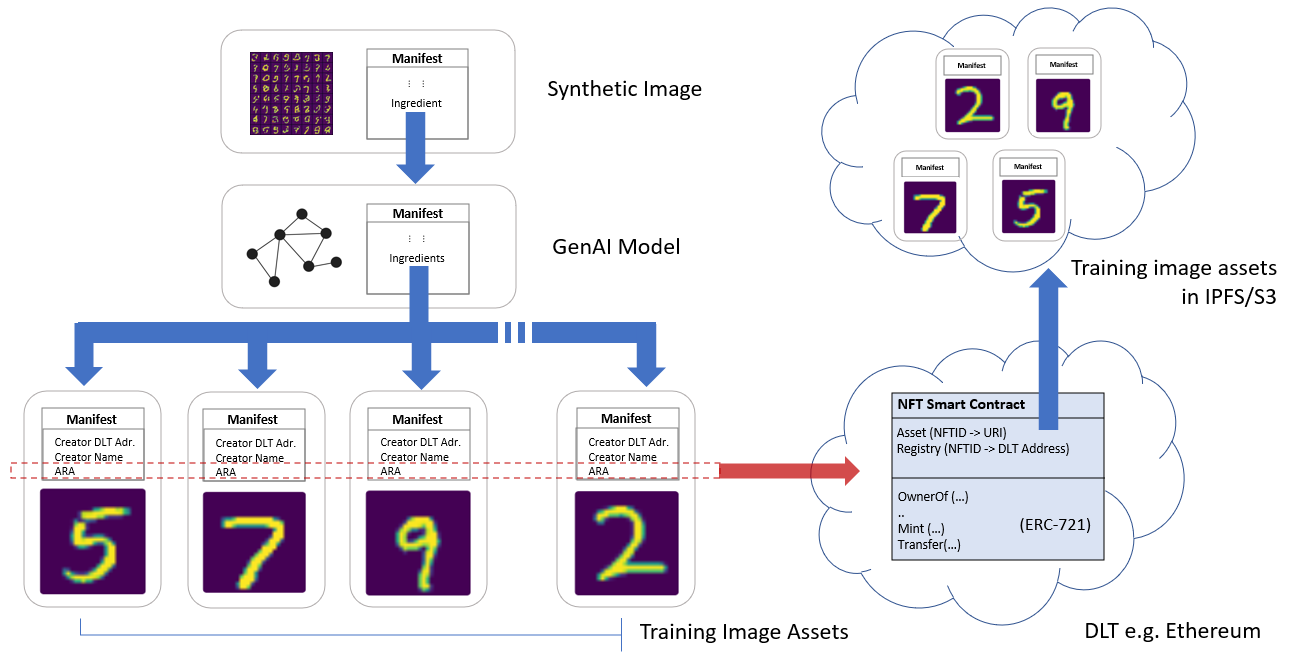}
    \caption{Left: Simple model.  The GAN-generated image bearing a C2PA manifest from which its provenance may be determined: the GenAI model (green) and the training data of that model (red). The Verify tool enables inspection of training data image manifests (blue), revealing the creator's name and DLT wallet address encoded in the metadata.  Right: Flexible model, integrating C2PA manifests with NFT to trace dynamic ownership of training data assets.  Training images are minted as NFTs and carry a link (ARA) to each NFT within the C2PA metadata of the asset.  The C2PA metadata thus provides a way to determine the current owner of an asset, including their wallet address, for receipt of royalties.  This obviates the need to store the wallet address statically in the ingredient manifests (per left). }
    \vspace{-1em}
    \label{fig:mnist}
\end{figure*}

\textbf{Content Attribution} has been explored from two perspectives: 1) detecting manipulation (\eg deep fakes \cite{kaggledf,mazaheri2022detection,zhao2021multi,zi2020wilddeepfake,guarnera2020deepfake}); 2) content attribution.  Attribution is the focus of cross-industry coalitions (\eg, CAI \cite{cai}, Origin \cite{origin}) and the emerging C2PA standard \cite{c2pa}, which we leverage. C2PA communicates provenance information \eg describing how an image was captured and what has been done to it, to aid users' trust decisions.  This is achieved by embedding `manifests' -- data packets -- within the image metadata.  The manifest encodes a signed graph structure of provenance records describing the content's capture and manipulation.  If the manifest is stripped from the asset, it may be recovered using a perceptual hash \cite{nguyen2021,Zhang2020manip,Bharati2021}, or watermark \cite{devi2009,baba2009,weng2019high} via look-up in a distributed database.  Recently, visual fingerprinting has been used to detect and attribute images to the GenAI models that made them \cite{yu2021responsible}.

\textbf{Distributed Ledger Technology (DLT)} (colloquially, `blockchain') enables multiple independent parties to share an append-only data structure (ledger) without a centralized point of trust \cite{dltgood}.  Beyond crypto-currencies, DLT has been used to track ownership of creative works via the ERC-721 Non-Fungible Token (NFT) standard \cite{erc721}.  NFTs are digital representations of assets, such as images, which may be openly traded on DLT, creating a history (provenance trail) of ownership. 
However, while owning an NFT grants the right to sell or dispose of that asset, it is undefined what else one has the right to do with the asset. EKILA adds a tokenized form of rights and leverages those smart contracts to mediate the payment of royalties to creators when those rights are exercised. Existing NFT royalty earning schemes are proprietary and limited to within-market sales, defeating the promise of NFT in creating decentralized markets for assets \eg  EIP-2981 \cite{eip2981} seeks to standardize royalty earning contracts for asset re-sales only. NFTs derive their value from scarcity but contain no counter-measures against copy-minting. We combine NFT with a creation provenance standard (C2PA) to mitigate this threat.


\section{EKILA Framework}

We first outline how C2PA may be applied to trace the provenance of a synthetic image to its training data  (subsec. ~\ref{subsec:c2pa}). We show how payment information, embedded immutably in those images at creation-time, may be used to reward contributors (subsec. ~\ref{subsec:mnist}).  We then extend that approach by describing how C2PA may be fused with NFT to trace {\em dynamic} ownership; \ie ownership that changes after asset creation (subsec. ~\ref{subsec:rights}).  Bridging {\em creation} provenance (C2PA) and {\em ownership} provenance (NFT) creates two sides of the triangular relationship (ORA) introduced in this paper, the third being the introduction of tokenized {\em rights}, described in subsec.~\ref{subsec:rights}.   This framework is combined in Sec.~\ref{sec:genai} with our visual attribution model to describe an end-to-end solution for synthetic image provenance.

\subsection{Recognizing and Rewarding Contribution}
\label{subsec:c2pa}

C2PA \cite{c2pa} manifests describe facts about the creation provenance of an asset, such as who made it, how, and which  `ingredient' assets were used in the process. These facts are called `assertions.'  Ingredients may point at assets, each bearing its manifest.  Thus C2PA encodes a graph structure rooted at the current asset, fanning out to its ingredients. Although C2PA manifests were initially developed with media assets (images, video, audio) in mind, any binary asset may bear a manifest. We, therefore, apply C2PA in our work to describe synthetic image provenance: to describe within an image's manifest the GenAI model used to produce it and, within a GenAI model, the ingredients used to train it.

\begin{figure*}[t!]
    \centering
    \includegraphics[width=0.87\linewidth,height=7.4cm]{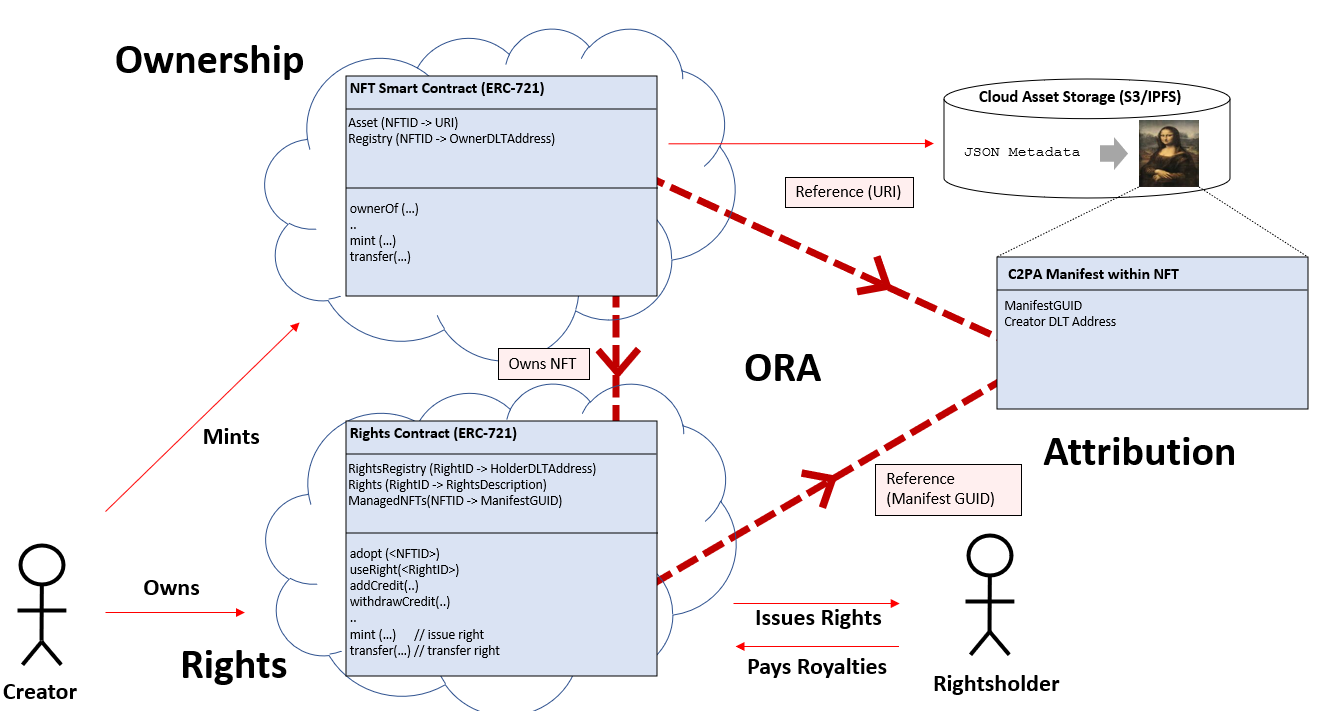}
    \caption{Ownership-Rights-Attribution (ORA) Triangle.  ORA ties together an NFT image and its ownership provenance (who owns/owned it), with its creation provenance (how it was created and what was done to it described via the C2PA attribution standard), and tokenized rights that may be granted by the creator to rightsholders.  ORA enables provenance to be traced from a synthetic media image to the GenAI model that made it and, ultimately, the training images responsible. The owners of those training images may then be recognized and rewarded for their contribution. }
    \vspace{-1em}
    \label{fig:ora}
\end{figure*}

\subsubsection{Contributor Recognition}
\label{subsec:mnist}

 C2PA provides for creator information to be asserted within a manifest; we include their name and crypto-currency (DLT) wallet address.  In Fig.~\ref{fig:mnist}, we show this is already sufficient to record GenAI contributors to a synthetic media item using a simple example of a  Generative Adversarial Network (GAN) model trained on the MNIST dataset\cite{mnist}.  MNIST is a dataset of 70K handwritten digits of 10 classes (digits 0-9). We embed C2PA manifests within all 60K images within the MNIST training partition, then train an unconditional GAN \cite{dcgan}.   A C2PA manifest is produced for the trained GAN model, referencing those training image ingredients.  A C2PA manifest is embedded within images generated by that GAN, with the sole ingredient being the GAN. Using a C2PA compliant open source SDK and tools, we can verify the provenance of the synthetic image given only its JPEG (Fig.~\ref{fig:mnist}, left): we determine its ingredient (GenAI model) and all the ingredients of that model (the training images).  Should a user of the synthetic image wish to recognize or reward the contributors of those training images, both the names and wallet addresses to make the transactions are obtained from the provenance graph.  Although in this example, we include training images individually in the manifest,  C2PA allows for manifests to be defined over archives (\eg zips)  of image collections for larger datasets (c.f. Sec.~\ref{sec:genai}).
 
 \subsubsection{Dynamic Ownership via NFT}
 \label{sec:nft}

The above approach is overly simplistic because asset ownership may change post-creation. NFTs provide a decentralized solution to record asset ownership of assets. NFTs exist as state within a DLT smart contract representing an asset collection. Under the dominant NFT standard (ERC-721), an NFT has a unique integer (NFTID) that maps to the wallet address of the current owner and also maps to a URI through which the asset may be accessed.  

We propose to bridge the creation provenance ecosystem described by C2PA with the ownership provenance ecosystem described by NFTs. C2PA provides the `Asset Reference Assertion' (ARA) to point at a location (URI) where the asset may be sourced.  Within the manifests of the training images, we create an ARA referencing the ingredient images, minted as NFTs, via an ad-hoc URI schema \texttt{c2pa-nft://dlt-id:address:nft-id}, which follows the draft CAIP-2 \cite{caip} proposal for describing a DLT smart contract address. 
For example, the URI \texttt{c2pa-nft://eip155:5:0x789/0x123} refers to the Ethereum Goerli test-net (\texttt{eip155:5}), with \texttt{0x789} the smart contract address of the relevant NFT collection on that DLT, and \texttt{0x123} the NFTID within that contract.  

\subsection{Tokenized Rights}
\label{subsec:rights}

EKILA introduces a way to define and assign usage rights to NFTs -- for example, the right to use an NFT image asset to train a GenAI model.  Neither NFT nor C2PA contains mechanisms for specifying rights.  We propose a smart contract --- the Rights contract -- to define rights and to distribute licenses to those rights, represented by {\em rights tokens}. Each creator manages a Rights contract, and issues tokenized rights to rightsholders via that contract.   Like NFTs, the Rights tokens subscribe to the ERC-721 standard (\eg with methods to transfer and inspect ownership) with additional methods related to rights and royalty payments.  The creator may mint rights to the asset as a separate step after minting the NFT (and its transfer to the Rights contract) is performed.  In this way, rights may be issued for using the asset in a decoupled way.   Discussion of the rights ontology is beyond the scope of this paper: any emerging digital rights description model could be utilized \cite{ccrel}. Creators may stipulate a royalty value to be paid when a license to a right is exercised (\eg to create a GenAI image), which we weigh according to a credit apportionment (subsec.~\ref{sec:visatt}).

\subsubsection{The ORA Triangle}

The relationship between the NFT, C2PA metadata in the asset, and the Rights contract encodes information on the Ownership, Rights and Attribution (ORA) of the asset, immutably bound in a triangular relationship (Fig.~\ref{fig:ora}):

\textbf{Ownership} The NFT encodes ownership and ownership history and references the asset via URI.  Note that ERC-721 does not guarantee the immutability of content at the URI, but the link to `Attribution' provides for this.  The Rights contract owns the NFT.

\textbf{Rights} The rights contract issues rights tokens owned by rightsholders.  It enables rights holders to pay the creator when exercising those rights.  The rights tokens embed the unique ID of the C2PA manifest to prevent the hijacking of the payment mechanism by rogue assets. 

\textbf{Attribution.} The C2PA manifest embedded in the asset expresses an immutable connection to the underlying content and its provenance history of creation, for example, any training data attributed to its creation: the manifest stores a unique identifier (GUID) and the wallet address of the creator.  The former is verifiable against the Rights contract, the latter against the NFT owner address.  The latter mitigates against the copyminting of the asset by an entity other than the creator.  The attribution of the asset to its training data is further developed via robust visual matching in Sec.~\ref{sec:visatt}.

ORA is applied to create an image asset in EKILA as follows: 1) The image is injected with a C2PA manifest describing its provenance; 2) An assertion is added to the manifest documenting the DLT address from which the asset will be minted; 3) The asset is minted to create an NFT \cite{erc721}; 4) Ownership of the NFT is transferred to the Rights contract operated by the creator; 5) The Rights contract stores the manifest identifier (GUID) of the NFT; 6) Rights may now be assigned (\eg sold) by the creator issuing tokens from the Rights contract.

\subsubsection{Exercising Rights and Royalties}

ORA provides a mechanism for a rightsholder to pay the asset creator via the creator's Rights contract.  In EKILA, the creator of a training image might receive a royalty payment when their work is used to train a GenAI model or when that model generates an image.  They issue a right to the GenAI model trainer, stipulating the base value of this award, which may subsequently be scaled according to the attribution of their training data to a generated image (c.f. \ref{sec:visatt}). The process starts with the C2PA manifest, tracing the provenance graph to identify the training images and, via the ARA, the associated NFTs (subsec.~\ref{sec:nft}).  The NFTs are owned by the Rights contracts of their respective creator, and their address may be obtained from the NFT (via the ERC-721 \texttt{ownerOf} method).  The Rights contract may then be interrogated to get the creator's wallet address and make a payment or donation.  EKILA also implements a stored value system within the Rights contract, enabling Rightsholders to pay into the smart contract and release funds to the creator as they exercise their rights. This enables ad-hoc payments (such as donations) and micro-payments without the high transaction cost of transferring crypto-currency. 

\begin{figure*}[t!]
    \centering
        \includegraphics[width=\linewidth,height=7cm]{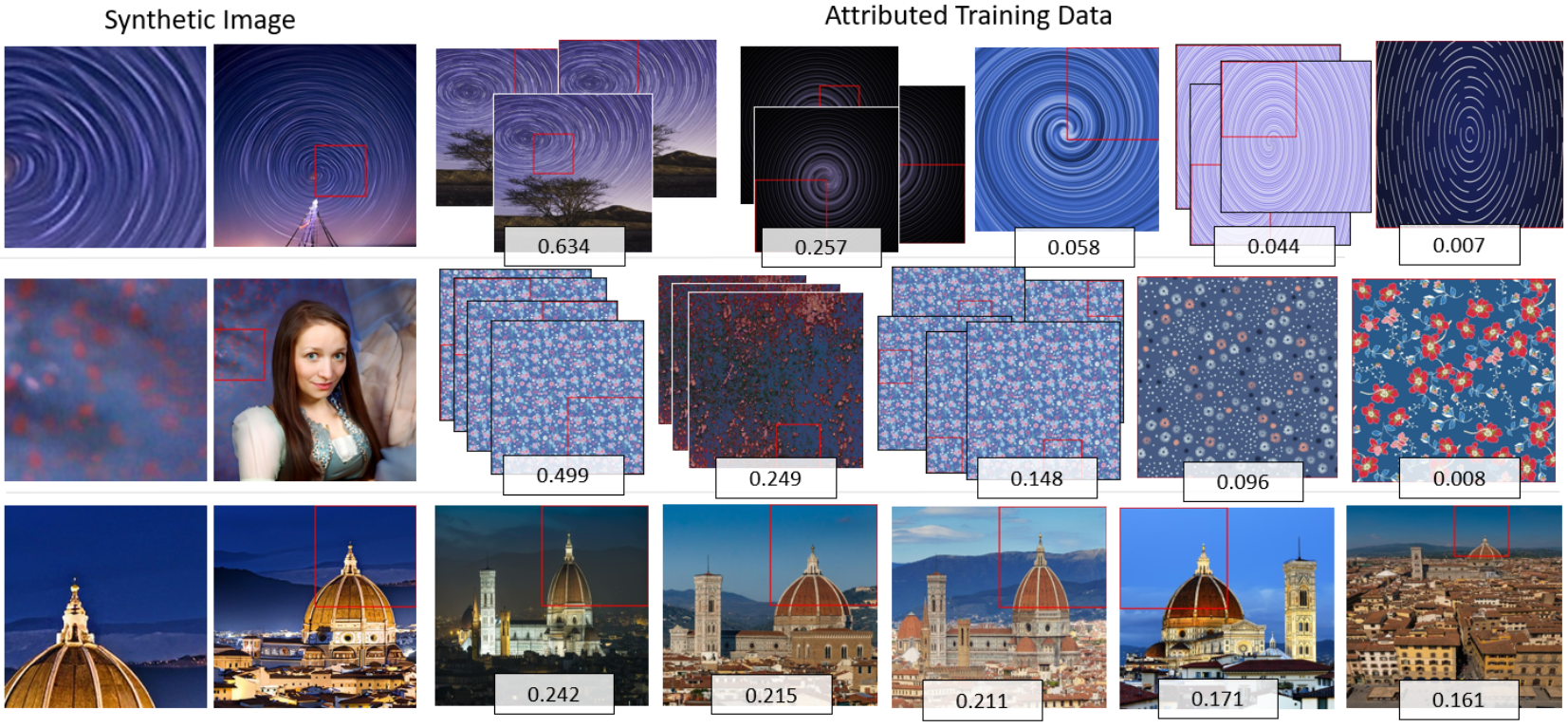}
    \caption{Attribution and apportionment over the IPF-Stock dataset showing examples of texture and object fragments attributed from three synthetic images.  Matching patches are outlined in red.  The prediction from the EKILA verifier (subsec.~\ref{sec:visatt}) is normalized across the top 5 contributing images (multiple contributions indicated as stacks in the figure) to form a credit apportionment to distribute to owners of each image via the ORA rights framework (subsec.~\ref{subsec:rights}) -- multiple patches attributed within a single training image accumulate their apportionment score.  Results are shown as re-ranked by apportionment (value inset). The final apportionment score forms a scaling factor on the royalty specified in the usage right(s) later distributed to the creator(s).}
    \vspace{-1em}
    \label{fig:attr-stock}
\end{figure*}

\begin{figure*}[t!]
    \centering
        \includegraphics[width=1.0\linewidth,height=4cm]{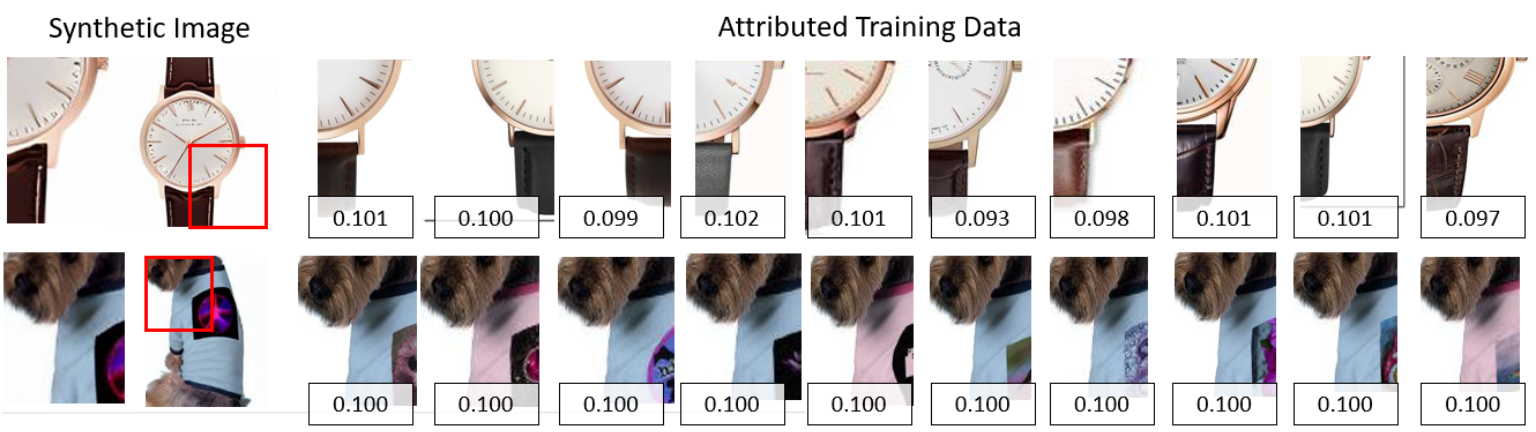}
    \caption{Attribution and apportionment over LAION showing several near-identical training images of products memorized by the GenAI and correctly attributed (top $K$=10 results shown). Even apportionment of credit reflects the similarity of the attributed examples.}
    \vspace{-1em}
    \label{fig:attr-laion}
\end{figure*}

\section{Visual Attribution and Apportionment}
\label{sec:genai}
 GenAI models\cite{ldm,imagen,dalle2} are typically trained on millions of images.  Recognizing and rewarding all contributors is impractical in this scenario.   Attributing credit to a subset of training data most correlated to the given synthetic image, and apportioning a weight across that subset, is necessary for a practical recognition and reward model.

\begin{table*}
\centering
\begin{adjustbox}{width=0.8\textwidth}

\begin{tabular}{ll|cccc|cccc}
\toprule
& & \multicolumn{4}{c} {\textbf{Image Attribution}} & \multicolumn{4}{c} {\textbf{Patch Attribution}}\\
\textbf{Dataset}  & \textbf{Method}                          & \textbf{R@1$\uparrow$} & \textbf{R@5$\uparrow$} & \textbf{R@10$\uparrow$} & \textbf{mAP$\uparrow$}   & \textbf{R@1$\uparrow$} & \textbf{R@5$\uparrow$} & \textbf{R@10$\uparrow$}  & \textbf{mAP$\uparrow$}  \\ 
\cmidrule{1-10}
\multirow{4}{*}{LAION} & EKILA (ours)           & \textbf{83.33}             &                    \textbf{54.31}                                           &            \textbf{52.94}               &    \textbf{59.26}                                                 & \textbf{87.50}               &                     \textbf{70.13}                                           &             \textbf{70.83}               &    \textbf{72.73}                                               \\ 

 & ViT-CLIP               &    56.25               &                          47.92                                      &               43.75                                                    &     47.61               &                     75.00                    &                               57.64                        &          56.02                                                                          &     59.91\\ 
 
 & RGB-P                                                                   &             5.88                       &          4.71                                                     &     3.26                                                    &     4.27               &                     12.6                                           &             3.32                        &          3.12                                                     &     4.38          \\ 
 
 & ALADIN               &    37.39               &                     33.74                           &             18.29                                                         &             28.87                                                                 &         \multicolumn{4}{c} {---}                                      \\ 
 
\cmidrule{1-10}
\multirow{4}{*}{IPF-Stock} & EKILA (ours)                                                             &  \textbf{70.83}               &                      \textbf{65.20}                                           &              \textbf{63.16}               &     \textbf{64.72}       &       \textbf{86.79}        &                            \textbf{67.36}              &                        \textbf{66.67}                         &    \textbf{70.03}                                          \\ 

 & ViT-CLIP               &    60.22               &                     58.60                                           &               51.61                                                    &     58.38               &                 67.74                    &                                62.9                        &          60.93                                                                          &     63.10              \\ 
 
 & RGB-P                                                                   &             3.58                        &          2.69                                                     &     0.27                                                    &     1.85               &                     19.4                                           &             14.8                        &          12.5                                                     &     15.1          \\ 
 
 & ALADIN               &    29.03               &                     23.23                           &             21.86                                                         &             23.50                                                                 &         \multicolumn{4}{c} {---}                                      \\

\bottomrule
\end{tabular}
\end{adjustbox}

\squeezeup
\caption{Attribution performance metrics for patch and whole image comparison using our proposed metric and three baselines (ViT-CLIP \cite{clip}, Patchwise RGB following \cite{carlini2023}, ALADIN \cite{aladin}) Comparing the performance for LDMs trained on LAION-400M and IPFree-Stock using   Recall @ $K=[1, 5, 10]$ and mAP, which rewards higher ranking of correct results up to $K=30$. }
\label{tab:patchresults}
\end{table*}

\subsection{Visual Attribution}\label{sec:visatt}

We consider visual attribution from the perspective of robust partial matching invariant to non-semantic changes in the content.
Our approach consists of three stages: 1) partial matching based on image fingerprints extracted at the image patch level, 2) pairwise verification and scoring of the most similar patch-level retrievals, and 3) distribution of credit based on the inferred patch similarity scores.  

\noindent \textbf{Patchified Fingerprinting.}
We adapt the whole-image visual fingerprinting approach outlined in \cite{Black_2021_CVPR} to enable large-scale retrieval of visually similar image patches.
As visual fingerprints, we consider compact embeddings of a CNN, contrastively trained to be discriminative of content whilst robust to degradations and manipulations.

Concretely, let $\phi_i=E(x_i)\in \mathbb{R}^{256}$ be the feature vector obtained as the output of a ResNet-50 encoder $E$ for an image patch $x_i$.
We train the patch encoder through a contrastive learning objective \cite{chen2020simple}
\begin{equation}
\mathcal{L}_{C} = - \sum_{i\in \mathcal{B}}   \log \left( \frac{d\left( \phi_i, \hat{\phi}_i \right)}{ d\left( \phi_i, \hat{\phi}_i \right) + \sum_{j \neq i \in \mathcal{B}} d\left( \phi_i, \phi_k \right)} \right),
\end{equation}
where $\hat{\phi}_i$ represents an embedding of a differently augmented version of $x_i$, $d(a, b):= \exp \left(\frac{1}{\lambda}  \frac{ {a}^\intercal { b}}{\Vert {a} \Vert_2 \Vert {b} \Vert_2} \right)$ measures the similarity between the feature vectors $a$ and $b$, and $\mathcal{B}$ is a large randomly sampled training mini-batch.
Besides the standard augmentation recipe used in contrastive learning (\eg, random cropping, color jittering, etc.), we consider content degradations due to noise, format change, and other manipulations as studied in \cite{hendrycks2019robustness}.

\noindent \textbf{Match Verification Model.}
Provided a shortlist of the top-$K$ candidate matches through the prior fingerprinting step, we now describe how we verify matches through an additional pair-wise comparison of the query with each candidate match.
This step is necessary since distance in the fingerprint embedding is generally insufficient to discriminate between true and false matches (c.f. Fig.~\ref{fig:roc}).
Rather than comparing images with global aggregated feature vectors (as in the scalable fingerprinting stage), we instead compare images at the level of spatial feature maps derived from that model. 
To this end, let $F_q\in \mathbb{R}^{H\times W \times D}$ be the feature map for a query patch $x_q$ and let similar $\{F_i\}_{i=1}^k$ be the $k$ corresponding retrieval feature maps.
We process each feature map with a $1\times1$ convolution to reduce the dimensionality to $\frac{D}{4}$ and then extract numerous pooled descriptors from a set of 2D feature map windows $\mathcal{W} \subset [1, H] \times [1, W]$, similar to R-MAC \cite{tolias2015particular}.
Let $f^q_{w}\in \mathbb{R}^\frac{D}{4}$ denote the GeM-pooled \cite{tolias2015particular} and unit-normalized feature vector for a window $w\in \mathcal{W}$ and feature map $F_q$.
In contrast to \cite{tolias2015particular}, we do not average these window-pooled feature vectors but collect them as:
\begin{equation}
    \hat{F}_q = [f^q_{w_1}, \ldots, f^q_{w_{|\mathcal{W}|}}] \in \mathbb{R}^{|\mathcal{W}| \times \frac{D}{4}},
\end{equation}
where $w_i \in \mathcal{W}$ and the number of windows is $|\mathcal{W}|=55$ in practice.
We then compute the feature correlation matrix 
\begin{equation}
    C_{qi} = \hat{F}_q \hat{F}^T_i \in \mathbb{R}^{|\mathcal{W}|\times |\mathcal{W}|}.
\end{equation}
These feature correlations are then flattened and fed to a 3-layer $\operatorname{MLP}$, which outputs a similarity score between query $q$ and retrieval $i$.
Concretely and to make the model symmetric w.r.t. its inputs, we define the match score between image patches $x_q$ and $x_i$ as 
\begin{equation}
    \operatorname{score}(x_q, x_i) = \sigma \big( \operatorname{MLP}(C_{qi}) + \operatorname{MLP}(C_{iq}) \big),
\end{equation}
where $\sigma$ represents a sigmoid activation. 

\noindent \textbf{Training the Verification Model.}
We build positive example pairs via data augmentation (like in the fingerprint pre-training) to train the match verification model and rely on hard-negative mining to produce challenging negatives.
We use a strong data augmentation protocol to generate positives, \ie, combinations of severe color jittering, blurring, random resize cropping, random rotations, etc.  
We maintain a queue $Q\in\mathbb{R}^{N \times \frac{D}{4}}$ of the most recent feature maps and assign each training query example its top-$K$ neighbors in $Q$ (in terms of cosine similarity of pooled features) as hard negative examples (we set $N=2^{14}$ and $K=20$ in practice). 
Given pairs of true and false matches, the model is trained with a standard binary cross-entropy loss.
During verifier training, we freeze the backbone feature extractor (the same as the fingerprint encoder $E$). 

\noindent \textbf{Apportioning Credit via Patch-Based Attribution.}
Finally, we describe how credit is assigned based on our patch-based attribution model.
Given an image $X_i \in \mathbb{R}^{H\times W \times 3}$, we represent it with the set of 21 patches $\mathcal{X}_i=\{x_1, \ldots, x_{21}\}$. We use all the $\frac{H}{2}\times \frac{W}{2}$ and $\frac{H}{4}\times \frac{W}{4}$ image tiles along with the whole image. 
For each image in the training set of the GenAI model, we extract patch fingerprints for all these patches. These patch embeddings, along with their position in the image and the image ID, are then stored in an inverted file index with Product Quantization (IVFPQ) \cite{faiss}, where large-scale approximate nearest-neighbor lookup can be performed efficiently. At query time, we extract the 21 patches $\mathcal{X}_q$ for a query image and assign a weight $w_{ij}$ to each image $\mathcal{X}_i$ in the database based on similarity to query patch $x_j \in \mathcal{X}_q$ via
\begin{equation}
    w_{ij} = \smashoperator{\sum_{x_k \in \mathcal{N}_K(x_j) \cap \mathcal{X}_i }} \max \big(\operatorname{score}(x_j, x_k)-\lambda, 0 \big), 
\end{equation}
where $\mathcal{N}_K(x_j)$ represent the top-$K$ retrieved patches from the database for query patch $x_j$ based on fingerprint similarity and $\lambda=0.7$ is a threshold for the verifier score.
We then assign credit per query patch by normalizing these weights over all images in the database. 
\begin{equation}
    \operatorname{credit}_{i}(x_j)=\frac{w_{ij}}{\sum_k w_{kj} }.
\end{equation}
This credit apportionment factor is illustrated in Figures~\ref{fig:attr-stock} and \ref{fig:attr-laion}.
Finally, credit for each image is summed over all query patches, resulting in a final attribution of $\operatorname{credit}_{i}=\sum_{x_j \in \mathcal{X}_q} \operatorname{credit}_{i}(x_j) $.  The weight is applied to the crypto-currency payments communicated to the owners of each image via the ORA framework.

\subsection{Datasets and Baselines}
\begin{figure}[t!]
    \centering
    \includegraphics[width=\linewidth,height=3.7cm]{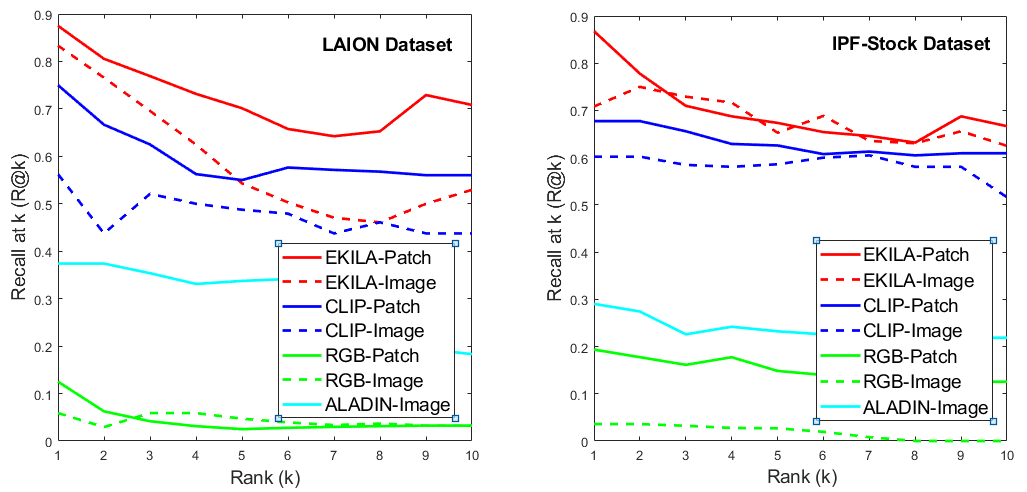}
    \caption{Performance (R@$K$) of our proposed method against common baselines for attribution (ViT-CLIP, RGB-P).  Vit-CLIP and ALADIN are semantic and style based representations.  EKILA outperforms all baselines across LAION and IPF-Stock.}
    \vspace{-1em}
    \label{fig:patk}
\end{figure}

\textbf{LAION-400M} \cite{laion400mdataset} is a dataset of 400M image-text pairs crawled from web pages between 2014-2021, and CLIP \cite{clip} filtered for consistency across modalities.  LAION is used extensively to train GenAI models, including Latent Diffusion Model (LDM) \cite{ldm}, which we use for our experiments.  LAION400M contains a large number of images depicting products and other intellectual property (IP) commensurate with the rights and attribution use case of EKILA. For our LAION experiments, we use a random 10M corpus for attribution, and sample product names from the 10M corpus to form 40 synthetic image queries.

\textbf{IPF-Stock} (IP Free) is a dataset of 200M public image assets and associated captions available on Adobe Stock.  Notably, the platform enforces strict content moderation to block products, brands, and other IP-encumbered visual content.  We make use of an LDM \cite{ldm} trained from scratch on this dataset, and sample 1M random images to form a corpus for attribution experiments. We sample captions from the 1M images to create 60 synthetic image queries.

\textbf{BAM-FG} \cite{aladin} is a dataset of 2.6M diverse digital artworks used to contrastively train our models (subsec~\ref{sec:visatt}).

\begin{figure}[t!]
    \centering
    \includegraphics[width=\linewidth,height=5.2cm]{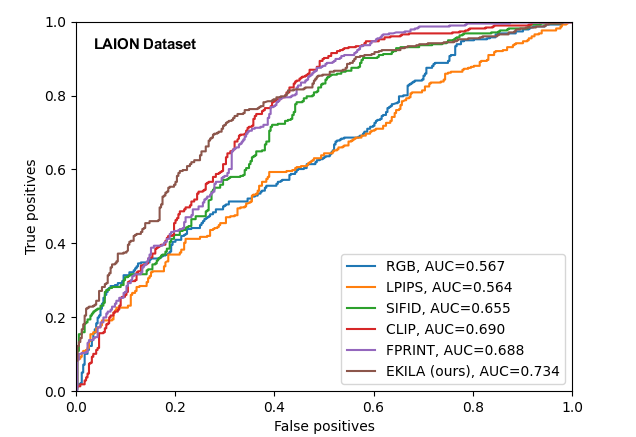}
    \caption{Verifying Attribution: ROC curve comparing performance of our proposed verifier to baseline embeddings and models at all thresholds over LAION (patch query set).}
    \vspace{-1em}
    \label{fig:roc}
\end{figure}

\textbf{Attribution Baselines.} We perform a large-scale search of each attribution corpus using our fingerprint embedding and verification models described in subsec.~\ref{sec:visatt}. We compare against the ViT-CLIP (CLIP with ViT image encoder \cite{clip}) semantic embedding and the ALADIN style-sensitive embedding. ViT-CLIP is used by emerging GenAI attribution tools \eg \cite{stableattribution}. We also compare against RGB values sampled over $4 \times 4$ non-overlapping image patches (RGB-P) proposed as a metric in a recent study of GenAI memorization \cite{carlini2023}. For pairwise verification, we additionally compare our proposed model against  LPIPS \cite{lpips} and SIFID \cite{sifid}.

\textbf{Metrics.} We measure attribution using: 1) Recall at $K$ ($R@K$);  the proportion of correctly attributed examples in the top $K= [1,5,10]$ results; 2) Mean average precision (mAP) which rewards a higher ranking of correct results.  For ground truth, we ran a perceptual study in which each query and ranked result pair are presented to Amazon Mechanical Turk (AMT) workers who decide if the pair are visually similar.

\subsection{Visual Attribution Experiments}

We first compare the performance of the end-to-end attribution search using the fingerprinting and verification model. 
As described in Sec.~\ref{sec:visatt}, an index of the attribution corpus over image patches is made, resulting in 210M patch embeddings for LAION and 21M patches for IPF-Stock. 
The experiment is replicated for ViT-CLIP, RGB-P, and ALADIN embeddings. 
ALADIN is designed to encode style only for whole images, as it performs poorly on image patches \cite{aladin}. The verification model is used to decide the top results for our approach using a decision threshold, determined empirically from the ROC experiment described shortly performed on a 100K  sample of held-out data.  Thresholds are applied similarly to the baseline embeddings.  
We present representative results of attribution on both LAION (Figs. 1,~\ref{fig:attr-laion}) and IPF-Stock (Fig.~\ref{fig:attr-stock}).

Table~\ref{tab:patchresults} shows that our fingerprinting plus verification method outperforms all baselines for both whole image queries and patch queries, outperforming ViT-CLIP by $~\sim 12\%$ mAP in both cases, with consistently higher performance (R@$K$) curves over all ranks $K=[1,10]$ (Fig.~\ref{fig:patk}) over both datasets, and for both partial and whole image matches.  Whilst attribution of visual style is important, ALADIN being explicitly focused upon style performs poorly relative to other baselines but may be substituted into EKILA to produce style attribution results (see sup.mat.).

We next evaluate the efficacy of the proposed verification model in deciding whether each image in the top $K=30$ results is an attribution or not.  Fig~\ref{fig:roc} analyses the performance of our pair-wise verification method versus other baselines for performing pair-wise comparison of images: thresholding ViT-CLIP and RGB-P embedding distance, and thresholding perceptual comparison models LPIPS and SIFID.  We compare on our largest dataset LAION. To compare without bias from threshold choice we perform Receiver-Operator-Curve (ROC) analysis (see Fig.~\ref{fig:roc}) for patch based queries; whole image queries performed similarly with AUC scores: EKILA 0.740, ViT-CLIP 0.699, FPRINT 0.657, SIFID 0.647, LPIPS 0.622, RGB 0.596.  FPRINT indicates simple thresholding of the fingerprint, \ie ablating the presence of the verifier and so justifies its inclusion in the pipeline.  The verifier outperforms the closest baseline of ViT-CLIP by over $4\%$ in both experiments.

\begin{figure}[t!]
    \centering
    \includegraphics[width=\linewidth,height=5.5cm]{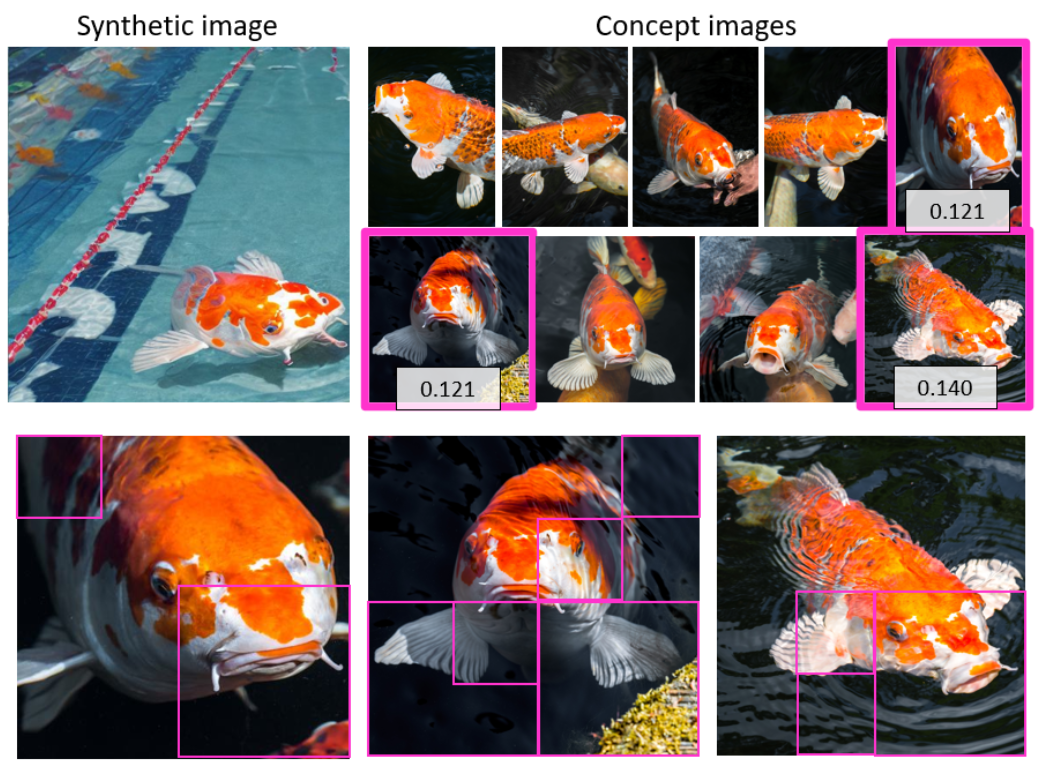}
    \caption{Dreambooth \cite{ruiz2022dreambooth} apportionment for ``Fish in a swimming pool" (top-left) fine-tuned using nine concept images of a specific fish species (top-right):  Bottom: High scoring patches within the three most highly apportioned concept images.}
    \vspace{-1em}
    \label{fig:dreambooth}
\end{figure}

\subsection{Apportionment experiments}

Figs.~\ref{fig:attr-stock}-\ref{fig:attr-laion} show how the prediction of the verifier, when normalized across the top results (subsec.~\ref{sec:visatt}), apportions (\ie linearly weighs) the reward provided to training data owners via ORA.  Multiple patches, often at multiple scales, may match a given synthetic image --- particularly in the absence of distinctive logos and products in IPF-Stock, this is due to distinctive texture patches matching.  We accumulate the apportionment scores per image, resulting in the weights shown in Fig.~\ref{fig:attr-stock}. Figs. 1, \ref{fig:attr-laion} show a  common case in LAION where many images exist of a given product leading to an apparent memorization of that product.  The apportionment is even across the similar training images in this case. 

Fig~\ref{fig:dreambooth} demonstrates an application for models fine-tuned via DreamBooth.  DreamBooth generates specific instances of an object given a prompt and a small number of `concept images' of that object \cite{ruiz2022dreambooth}.  The concept images are treated as a small attribution corpus, for which we mint NFTs via the ORA framework.  The apportionment accumulates prediction scores from the verifier for each concept image, and payments are made. 

\section{Conclusion}

We presented EKILA; a framework for recognizing and rewarding creative contributions to GenAI. We first described how the C2PA standard may be applied for data provenance in GenAI linked to the NFT ecosystem to describe content ownership.  Building on a small-scale demonstration of this concept using MNIST, we showed how NFT may be extended to provide a mechanism for dynamically issuing rights and receiving royalties.  We introduced a robust visual attribution method to assign credit to a relevant subset of GenAI training data for a synthetic image.  We showed the model to outperform emerging approaches for attribution based on ViT-CLIP, and how our model may apportion credit over the attributed training data. We demonstrated this for LDMs trained on two large datasets (LAION, IPF-Stock)  and showed a further use case in apportioning credit over Dreambooth concept images. 

Like emerging papers \cite{somepalli2022, carlini2023} and tools \cite{stableattribution} addressing content attribution, we assume correlation between similarity and attribution. Causal relationships for attribution such as `leave one out' model re-training are common for small-scale datasets but impractical for the scale of datasets used in GenAI. Our assumption is therefore that correlation is a good approximation to causation. In the future, the fusion of correlation via large-scale similarity with causal approaches e.g. over a matched shortlist would be a promising direction to develop GenAI attribution. Future work on ORA should study the socio-economic drivers necessary to make our prototype sustainable in the wild \eg legal topics on digital rights and digital rights description languages. Many of these questions may remain open for some time, yet ensuring fair attribution for creatives in GenAI is both urgent and timely. We believe EKILA presents a promising step toward an end-to-end solution to address this problem.

\section*{Acknowledgement}

Implementation of the EKILA prototype was supported in part by DECaDE under EPSRC Grant EP/T022485/1. 

{\small
\bibliographystyle{ieee_fullname}
\bibliography{ekila}
}

\end{document}